\documentclass{SCIS2019}

\usepackage{wrapfig}
\usepackage{booktabs}
\usepackage{algorithm}
\usepackage{algorithmic}
\usepackage{color}
\usepackage{xcolor}
\usepackage{bm}
\usepackage{multirow}
\usepackage{bbding}
\usepackage{bbm}
\DeclareMathOperator*{\argmax}{arg\,max}
\makeatletter
\newcommand\figcaption{\def\@captype{figure}\caption}
\newcommand\tabcaption{\def\@captype{table}\caption}
\makeatother

\begin{document}
\ArticleType{RESEARCH PAPER}
\Year{2019}
\Month{}
\Vol{}
\No{}
\DOI{}
\ArtNo{}
\ReceiveDate{}
\ReviseDate{}
\AcceptDate{}
\OnlineDate{}

\title{Sparse spatial transformers for few-shot learning}{Sparse spatial transformers for few-shot learning}

\author[]{Haoxing CHEN}{}
\author[]{Huaxiong LI}{{huaxiongli@nju.edu.cn}}
\author[]{Yaohui LI}{}
\author[]{Chunlin CHEN}{}

\AuthorMark{Haoxing Chen}

\AuthorCitation{Chen H X, Li H X, Li Y H, et al}


\address[]{Department of Control Science and Intelligence Engineering, Nanjing University, Nanjing {\rm 210093}, China}

\abstract{Learning from limited data is challenging because data scarcity leads to a poor generalization of the trained model. A classical global pooled representation will probably lose useful local information. Many few-shot learning methods have recently addressed this challenge using deep descriptors and learning a pixel-level metric. However, using deep descriptors as feature representations may lose image contextual information. Moreover, most of these methods independently address each class in the support set, which cannot sufficiently use discriminative information and task-specific embeddings. In this paper, we propose a novel transformer-based neural network architecture called sparse spatial transformers (SSFormers), which finds task-relevant features and suppresses task-irrelevant features. Particularly, we first divide each input image into several image patches of different sizes to obtain dense local features. These features retain contextual information while expressing local information. Then, a sparse spatial transformer layer is proposed to find spatial correspondence between the query image and the full support set to select task-relevant image patches and suppress task-irrelevant image patches. Finally, we propose using an image patch-matching module to calculate the distance between dense local representations, thus determining which category the query image belongs to in the support set. Extensive experiments on popular few-shot learning benchmarks demonstrate the superiority of our method over state-of-the-art methods. Our source code is available at \url{https://github.com/chenhaoxing/ssformers}.}

\keywords{few-shot Learning, transformer, metric-learning, cross-attention}

\maketitle

\section{Introduction}
With the availability of large-scale labeled data, visual understanding technology has made substantial progress in many tasks~\cite{1,2}. However, collecting and labeling such a large amount of data is time-consuming and laborious. Few-shot learning is committed to solving this problem, enabling deep models to have better generalization ability even on a few samples~\cite{41,42,43}.

Many few-shot learning methods have recently been proposed and can be roughly divided into two categories: optimation~\cite{3,4,5} and metric-learning~\cite{6,7,8,9,10}-based meta-learning methods. Optimation-based methods aim to learn transferable meta-knowledge to handle new tasks in the process of learning multiple tasks~\cite{3}. Metric-learning-based methods focus on learning a good feature representation or distance metric~\cite{7,10}. These methods have attracted widespread attention due to their simplicity and effectiveness. Therefore, we mainly focus on metric-learning-based methods in this paper.

\begin{figure}[t]
	\centering
	\includegraphics[width=16cm, height=8.2cm]{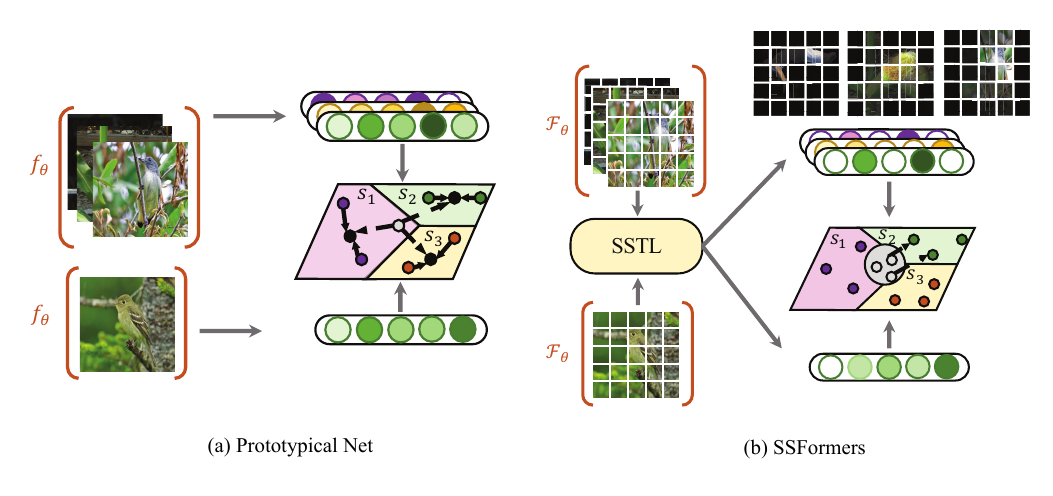} 
	\caption{Prototypical Net~\cite{7} learns a global-level representation in an appropriate feature space and uses Euclidean distance to measure similarities. In contrast, our model first generates dense local representations through image patches and then uses the sparse spatial transformer layer (SSTL) to select task-relevant patches, generating task-specific prototypes. Finally, similarities are obtained by matching between attentioned image patches.}
\end{figure}
For feature representations, most of the existing metric-learning based methods~\cite{6,7,11} adopt global features for recognition, which may cause helpful local information to be lost and overwhelmed.
Recently, DN4~\cite{9}, MATANet~\cite{12}, and DeepEMD~\cite{13} adopted dense feature representations (i.e., deep descriptors) for few-shot learning tasks, which are more expressive and compelling than using global features. Another branch that enhances image representation uses the attention mechanism to align the query image with the support set. For example, crossattention networks (CANs)~\cite{14} and SAML~\cite{15} use the semantic correlation between the support set and query image to highlight the target object.

For distance metrics, existing dense feature-based methods usually adopt a pixel-level metric, and the query image is taken as a set of deep descriptors. For example, in DN4~\cite{9}, for each deep query descriptor, these methods find its nearest neighbor descriptors in each support class. Additionally, CovaMNet~\cite{16} calculates a local similarity between each query deep descriptor and a support class using a covariance metric.

However, most existing methods use global features or deep descriptors, which are ineffective for few-shot image recognition. Because global features lose local information, deep descriptors lose the contextual information of images. Moreover, the above methods independently process each support class and cannot use the context information of the entire task to generate task-specific features.

This paper proposes a novel transformer-based architecture for few-shot learning called sparse spatial transformers (SSFormers). SSFormers extract the spatial correlation between the query image and the current task (the entire support set) to align task-relevant image patches and suppress task-irrelevant image patches. As shown in Figure 1, we first divide each input image into several patches and obtain dense local features. Second, we select task-relevant query patches by a function of mutual feeling, i.e., the mutual nearest neighbor~\cite{17,18}. The selected query patches are then used to generate task-specific prototypes. Finally, a patch-matching module (PMM) is proposed to measure the similarity between query images and aligned support classes. For each patch from a query image, the PMM calculates its similarity scores to the nearest neighbor patch in each aligned class prototype. Then, similarity scores from all query patches are accumulated as a patch-to-class similarity.

The main contributions of this work are summarized as follows: 

\begin{itemize}
	\itemindent 2.8em
	\item[(1)] We propose a novel \emph{sparse spatial transformers} for few-shot learning, which can select task-relevant patches and generate a task-specific prototype.
	
	\item[(2)] We propose a \emph{patch-matching module} to obtain similarity between query images and task-specific prototypes. Experiments prove that this approach is more suitable for image patch-based feature representation than directly using the cosine similarity.
	
	\item[(3)] We conduct extensive experiments on popular few-shot learning benchmarks and show that the proposed model achieves competitive results compared to other state-of-the-art methods.
\end{itemize}

\section{Related Work}
Few-shot recognition aims to learn transferable meta knowledge from seen classes. In general, most representative few-shot learning methods based on meta-learning can be organized into two categories as follows.
\subsection{Optimation-based Meta-learning Methods}
Optimation-based methods focus on training models that can perform well in unseen tasks with only a few fine-tuning steps.
Finn \textit{et al.}~\cite{3} proposed a model-agnostic meta-learning algorithm that learns the sensitive initial parameters of the networks to generalize to new tasks with several iterations. Ravi \textit{et al.}~\cite{35} proposed a long short-term memory-based meta-learner, which ensures that initialization allows the fine-tuning step to start at an appropriate point. Because the initial model of the ordinary meta-learning-based method could be too biased toward existing tasks to adapt to new tasks, Jamal \textit{et al.}~\cite{36} proposed an entropy-based method that can learn an unbiased initial model. Instead of forcibly sharing an initialization between tasks, Li \textit{et al.}~\cite{37} learned to generate matching networks by learning transferable meta-knowledge knowledge across tasks and directly producing network parameters for similar unseen tasks. However, how to reduce the computational cost and ensure the diversity and validity of synthetic data are considerable challenges for the meta-learning-based method.

\subsection{Metric-learning-based Meta-learning Methods}
\textbf{Global feature-based methods.} The traditional metric-learning-based few-shot learning methods use an additional global average pooling layer to obtain the global feature representation at the end of the backbone and use different metrics for recognition.
MatchingNet~\cite{6} uses the cosine distance to measure the similarity between the query image and each support class.
Prototypical Net~\cite{7} takes the empirical mean as the prototype representation of each category and uses Euclidean distance as the distance metric. RelationNet~\cite{8} proposes a nonlinear learnable distance metric. These methods based on global features lose much useful local information, which harms recognition tasks under few-shot learning settings.

\textbf{Dense feature-based methods.} Another metric-learning-based method branch uses pixel-level deep descriptors as feature representations. DN4~\cite{9} uses the $k$-nearest neighbor algorithm to obtain the pixel-level similarity between images. MATANet~\cite{12} proposes a multiscale task adaptive network to select task-relevant deep descriptors at multiple scales. DeepEMD~\cite{13} proposes a differentiable earth mover's distance to calculate the similarity between image patches. Our SSFormers also belong to this method based on dense features. An important difference in our method is that we divide input images into several patches of different sizes and extract features. Compared with global features, the features extracted by our method can express local information. Furthermore, compared with deep descriptors, the extracted features contain context information.

\textbf{Attention-based methods.} CANs~\cite{14} propose a crossattention algorithm to highlight the common objects in an image pair. SAML~\cite{15} proposes a collect-and-select strategy to align the main objects in an image pair. RENet~\cite{10} improves network generalization performance over unseen categories from a relational perspective. Differently, our SSFormers select task-relevant patches in the query image to align the support set with the query image by a sparse spatial crossattention algorithm.

\textbf{Transformer-based methods.} FEAT~\cite{19} first introduces transformer~\cite{20} to few-shot learning. FEAT uses a transformer to conduct support set sample relationships and generate task-specific support features. CrossTransformers~\cite{21} proposes to use a self-supervised learning algorithm to enhance the feature representation ability of the pretrained backbone and use a transformer to achieve alignment.

\begin{figure*}[t]
	\centering
	\includegraphics[width=16cm, height=11cm]{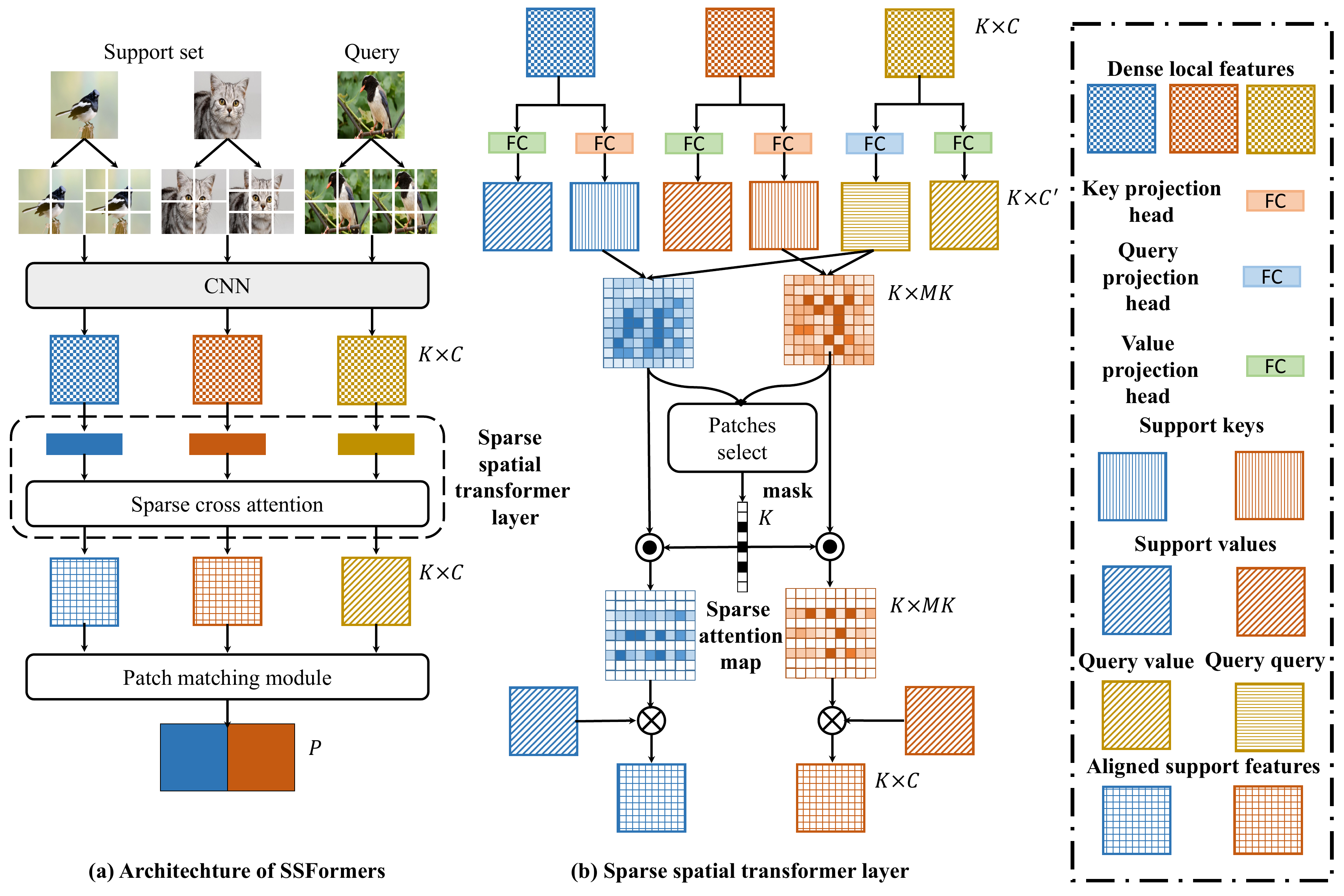} 
	\caption{Illustration of the proposed SSFormers. We propose to generate dense local features and find task-relevant features through a sparse spatial transformer layer.}
	\label{fig2}
\end{figure*}

\section{Preliminary}
We first introduce the problem definition of few-shot learning.
Few-shot learning is dedicated to learning transferable knowledge between tasks and using the learned knowledge to solve new tasks. In the few-shot learning scenario, the task is usually set as \emph{N}-way \emph{M}-shot, where \emph{N} is the number of categories, and \emph{M} is the number of labeled samples in each category. Under this setting, the model is trained on a training set $\mathcal{D}_{train}$ with a large amount of labeled data. We use episodic training mechanisms to train our model to learn transferable knowledge. For training, the episodic training mechanism samples batched tasks from $\mathcal{D}_{train}$. In each episode, we first construct query set $\mathcal{D}_{Q}=\{ (x^q_i,y^q_i) \}_{i=1}^{N \times B}$ and support set $\mathcal{D}_S=\{ (x^s_i,y^s_i) \}_{i=1}^{N\times M}$, where $B$ is a hyperparameter that we must fix in our experiments. Typically, $B$ is set to 15~\cite{14,19}. Then, our model predicts to which support set category each sample in the query set belongs. When the model training is completed, we sample tasks from unlabeled test sets $\mathcal{D}_{test}$ to verify the model's performance.

\section{Our Method}
In this section, we first introduce our method for generating dense local representations. Then, we describe our sparse spatial transformers layer, which spatially aligns query images and support classes. Finally, we describe the PMM used to calculate the final similarities. An overview of our framework is shown in Figure 2.

\subsection{Dense Local Feature Extractor}
The metric-learning-based few-shot learning method aims to find an effective feature representation and a good distance metric to calculate the similarity between images. In contrast to the methods that use global features, deep descriptor-based methods have achieved better results because deep descriptors contain richer and more transferable semantic information. However, these methods based on deep descriptors lose the contextual information of images. To combine the advantages of the two branches of methods and reduce the disadvantages, our SSFormers aim to establish hierarchical local representations for spatial comparison. 

As illustrated in Figure 2, dense local representations extractor $F_{\theta}$ evenly divides the image into $H\times W$ patches, and the backbone network individually encodes each image patch to generate a feature vector. The feature vectors generated by all patches constitute each image's dense local representations set. Because single-scale patches may not be able to fully obtain the local details and context information of the image, we adopt a pyramid structure to generate hierarchical local representations. Thus, the feature representation of an input image $x$ can be denoted as $F_{\theta}(x) \in \mathbb{R}^{K \times C}$, where $C$ is the number of channels, and $K$ is the number of all local patch representations. Particularly, we adopt two image patch division strategies of size 2$\times$2 and 4$\times$4 to obtain 20 dense local representations.

In each \emph{N}-way \emph{M}-shot few-shot image recognition task, for each support class, we have $M$ samples and obtain \emph{M} feature representations. Instead of using the empirical mean of \emph{M} feature representations~\cite{7} to obtain the class representation, we use all the patches in each support class, i.e., $S_n\in\mathbb{R}^{MK\times C}$,
where $S_n$ is the class representation of the $n$-th support class. The entire support set representation can be denoted as $S\in \mathbb{R}^{N\times MK \times C}$. Similarly, for a query image $x_i^q$, through $F_{\theta}$, we can obtain feature representation $q = F_{\theta}(x_i^q)\in\mathbb{R}^{K \times C}$, where $i=\{1,...,BN\}$.

\subsection{Sparse Spatial Transformers Layer}
Sparse spatial transformers aim to enhance the discriminant ability of local feature representations by modeling the interdependencies between different patches in the query image and the entire support set. In an \emph{N}-way \emph{M}-shot task, key $k_S$ and value $v_S$ are generated for support set feature $S$ using two independent linear projections: the key projection head $h_k$: $\mathbb{R}^C \mapsto \mathbb{R}^{C'}$ and the value projection head $h_v$: $\mathbb{R}^C \mapsto \mathbb{R}^{C'}$. Similarly, the query image feature $q$ is embedded using the value projection head $h_v$ and the query projection head $h_q$: $\mathbb{R}^C \mapsto \mathbb{R}^{C'}$ to obtain the value $v_q$ and query $q_q$.

Inspired by~\cite{17}, which proposed the MNN algorithm to eliminate batch effects in single-cell RNA sequencing data, we argue that if patch $q_i$ feels that $S_j$ is its closest patch and vice versa, then they are likely to have similar local features, where $q_i\in q_q, i\in\{1, ..., K\}$ and $S_j\in k_S, j\in\{1, ..., NMK\}$. Conversely, if patch $S_j$ feels that $q_i$ is not such a close patch, then even if $q_i$ feels that $S_j$ is its closest patch, the actual relationship between them is relatively weak. In other words, the correlation between two patches is a function of mutual feeling, not a one-way feeling. Therefore, we can use this bidirectionality to select task-relevant patches in the current task.

We first calculate the semantic relation matrix between the query image and support class $n$ and obtain $\mathbf{R}_n$:

We first calculate the semantic relation matrix between the query image and support class $n$, and get $\mathbf{R}_n$:
\begin{equation}
	\mathbf{R}_n = \frac{q_q\times k_{S_n}^\top}{\sqrt{C'}}\in \mathbb{R}^{K \times MK}.
\end{equation}

To find task-relevant patches, we concatenate all semantic relation matrixes $\mathbf{R}_n, n=\{1, ..., N\}$ to obtain $\mathbf{R}\in \mathbb{R}^{K \times NMK}$. Each row in $\mathbf{R}$ represents the semantic similarity of each patch in the query image to all patches of all images in the support set.

Particularly, we propose a novel sparse spatial crossattention algorithm to find task-relevant patches in the query image.
For each patch $q_i\in q_q$, we find its nearest neighbor $n_q^i$ in $k_S$, and then find the nearest neighbor $n_S^i$ of $n_q^i$ in $q_q$. If $i=n_S^i$, then we consider $q_i$ to be a task-relevant patch. After collecting all task-relevant patches in $q_q$, we can get the sparse mask $m=[m_1;...;m_K]$, which can be computed as:
\begin{gather}
	n_q^i = {\argmax_{j} }\mathbf{R}_{i,j},\\
	n_S^i = \argmax_{k} \mathbf{R}_{k,n_q^i},\\	
	m^i = \mathbbm{1}(i = n_S^i),
\end{gather}
where $\mathbbm{1}$ is the indicator function: When $i = n_S^i$, $\mathbbm{1}$ is equal to 1; otherwise, it is 0. Using sparse mask $m$ and semantic relation matrix $\mathbf{R}_n$, we can obtain sparse attention map $a^n$ and use it to align each support class $n$ to query image $q$ and obtain task-specific prototype $v_{S_n|q}$, which can be computed as:
\begin{gather}
	a_n = m * \mathbf{R_n} \in \mathbb{R}^{K\times MK}, \\
	v_{S_n|q} = a_n\times v_{S_n}^\top\ \in \mathbb{R}^{K\times C'}.
\end{gather}

\subsection{Patch-matching Module}
The PMM is built as a similarity metric with no parameter to train. Given query value $v_q\in \mathbb{R}^{K\times C'}$ and the aligned prototype of class $n$ $v_{S_n|q} \in \mathbb{R}^{K\times C'}$, we can obtain their patch-to-patch similarity matrix as follows:
\begin{gather}
	\mathbf{D}^n = \frac{v_q\times v_{S_n|q}}{||v_q||\cdot||v_{S_n|q}||} \in \mathbb{R}^{K\times K}.
\end{gather} 
Then, for each patch in $v_q$, we select the most similar patch among all patches from prototype $v_{S_n|q}$. We sum $K$ selected patches as the similarity between the query image and support class $n$:
\begin{gather}
	P^n = \sum_{i=1}^{K}  \max\limits_{j\in \{1,...,K\}} \mathbf{D}^n_{i,j}.
\end{gather}
Under the \emph{N}-way \emph{M}-shot few-shot learning setting, we can obtain semantic similarity vectors $P\in \mathbb{R}^N$. The training procedure of SSFormers is shown in Algorithm 1.

\begin{algorithm}[t]
	\caption{Training strategy of SSFormers}
	\textbf{Require:} Training set $\mathcal{D}_{train}$
	\begin{algorithmic}[1] 
		\FORALL{iteration=1, ..., MaxIteration}
		\STATE Sample $N$-way $M$-shot task $(\mathcal{D}_{Q},\mathcal{D}_S)$ from $\mathcal{D}_{train}$
		\STATE Compute $S= F_{\theta}(\mathcal{D}_S) \in \mathbb{R}^{N\times MK \times C}$
		\FOR{$i$ in $\{1,...,NB\}$}
		\STATE Compute $q = F_{\theta}(x_i^q)\in\mathbb{R}^{K \times C}$
		\STATE Compute sparse attention map using Eqs. (1)?(5)
		\STATE Obtain task-specific prototype $v_{S_n|q}$ using Eq. (6)
		\STATE Compute similarity $P_i$ using Eqs. (7) and (8)
		\ENDFOR
		\STATE Obtain cross-entropy loss $\mathcal{L}=\sum_{i=1}^{NB}{\rm CE}(P_i, y^q_i)$
		\STATE Update parameters in SSFormers by SGD
		\ENDFOR
		\RETURN Trained SSFormers
	\end{algorithmic}
\end{algorithm}

\begin{table}[t]
	\centering
	\caption{Average recognition accuracy of 5-way 1-shot and 5-way 5-shot tasks with $95\%$ confidence intervals on \emph{mini}ImageNet and \emph{tiered}ImageNet. $^\dagger$ denotes that it is our reimplementation under the same setting. (The top two performances are shown in bold.)}
	\begin{tabular*}{\hsize}{@{}@{\extracolsep{\fill}}cccccc@{}}
		\toprule
		\label{mini}
		\multirow{2}{*}{Method}  &\multirow{2}{*}{Backbone} & \multicolumn{2}{c}{\emph{mini}ImageNet}& \multicolumn{2}{c}{\emph{tiered}ImageNet}\\
		\cline{3-4}		\cline{5-6}
		&&5-way 1-shot & 5-way 5-shot&5-way 1-shot & 5-way 5-shot\\
		\midrule
		Prototypical Net \cite{7} & Conv-64F & 49.42$\pm$\footnotesize{0.78} & 68.20$\pm$\footnotesize{0.66}&53.31$\pm$\footnotesize{0.89} & 72.69$\pm$\footnotesize{0.74} \\
		CovaMNet \cite{16} &Conv-64F& 51.19$\pm$\footnotesize{0.76} &67.65$\pm$\footnotesize{0.63}&54.98$\pm$\footnotesize{0.90} & 71.51$\pm$\footnotesize{0.75}\\
		DN4 \cite{9} & Conv-64F  & 51.24$\pm$\footnotesize{0.74} &\textbf{71.02$\pm$\footnotesize{0.64}} & 53.37$\pm$\footnotesize{0.86} &\textbf{74.45$\pm$\footnotesize{0.70}} \\
		SAML \cite{15} &Conv-64F & 52.22$\pm$\footnotesize{0.00}& 66.49$\pm$\footnotesize{0.00}&-&- \\
		DSN \cite{25} & Conv-64F &  51.78$\pm$\footnotesize{0.96}& 68.99$\pm$\footnotesize{0.69}& 53.22$\pm$\footnotesize{0.66}& 71.06$\pm$\footnotesize{0.55}  \\	
		DeepEMD$^\dagger$ \cite{13} & Conv-64F  & 52.15$\pm$\footnotesize{0.28}& 65.52$\pm$\footnotesize{0.72}& 50.89$\pm$\footnotesize{0.30}& 66.12$\pm$\footnotesize{0.78} \\
		CFMN \cite{33} & Conv-64F& \textbf{52.98$\pm$\footnotesize{0.84}}& 68.33$\pm$\footnotesize{0.70}& -& -\\
		CTX$^\dagger$ \cite{21}  & Conv-64F & 52.38$\pm$\footnotesize{0.20}& 68.34$\pm$\footnotesize{0.16} &\textbf{55.32$\pm$\footnotesize{0.22}} &73.12$\pm$\footnotesize{0.19}\\
		\midrule
		SSFormers &Conv-64F  &\textbf{55.00$\pm$\footnotesize{0.22}}  & \textbf{70.55$\pm$\footnotesize{0.17}}&\textbf{55.54$\pm$\footnotesize{0.19}}  & \textbf{73.72$\pm$\footnotesize{0.21}}  \\
		\midrule
		Prototypical Net \cite{7} & ResNet12 & 62.59$\pm$\footnotesize{0.85} & 78.60$\pm$\footnotesize{0.16}& 68.37$\pm$\footnotesize{0.23} & 83.43$\pm$\footnotesize{0.16} \\
		CAN \cite{14} &  ResNet12 &  63.85$\pm$\footnotesize{0.48}& 79.44$\pm$\footnotesize{0.34} & 69.89$\pm$\footnotesize{0.51}& 84.23$\pm$\footnotesize{0.37} \\
		DSN \cite{25} & ResNet12 &  62.64$\pm$\footnotesize{0.66}& 78.83$\pm$\footnotesize{0.45} & 67.39$\pm$\footnotesize{0.82}& 82.85$\pm$\footnotesize{0.56} \\	
		DeepEMD \cite{13} &  ResNet12 &  65.91$\pm$\footnotesize{0.82}& 82.41$\pm$\footnotesize{0.56} & 71.16$\pm$\footnotesize{0.87}& 83.95$\pm$\footnotesize{0.58}\\		
		FEAT \cite{19} &  ResNet12 &  66.78$\pm$\footnotesize{0.20}& 82.05$\pm$\footnotesize{0.14}& 70.80$\pm$\footnotesize{0.23}& 84.79$\pm$\footnotesize{0.16} \\
		gLoFA  \cite{26} &  ResNet12&  66.12$\pm$\footnotesize{0.42}& 81.37$\pm$\footnotesize{0.33} & 69.75$\pm$\footnotesize{0.33}& 83.58$\pm$\footnotesize{0.42} \\
		ArL \cite{27} &  ResNet12 & 65.21$\pm$\footnotesize{0.58}& 80.41$\pm$\footnotesize{0.49} &-&-\\	
		PSST \cite{28} &  ResNet12 &  64.05$\pm$\footnotesize{0.49}& 80.24$\pm$\footnotesize{0.45} &-&-\\
		RENet \cite{10} & ResNet12 &  \textbf{67.60$\pm$\footnotesize{0.44}} & \textbf{82.58$\pm$\footnotesize{0.30}} & \textbf{71.61$\pm$\footnotesize{0.51}}& \textbf{85.28$\pm$\footnotesize{0.35}} \\
		NCA(Nearest Centroid) \cite{34} & ResNet12 & 62.55$\pm$\footnotesize{0.12}& 78.27$\pm$\footnotesize{0.09} &68.35$\pm$\footnotesize{0.13} &83.20$\pm$\footnotesize{0.10}\\
		CTX$^\dagger$  \cite{21} &ResNet12 & 63.95$\pm$\footnotesize{0.21}& 79.54$\pm$\footnotesize{0.17} &70.22$\pm$\footnotesize{0.11} &83.78$\pm$\footnotesize{0.16}\\
		\midrule		
		SSFormers &ResNet12  &\textbf{67.25$\pm$\footnotesize{0.24}}  & \textbf{82.75$\pm$\footnotesize{0.20}}  &\textbf{72.52$\pm$\footnotesize{0.25}}  & \textbf{86.61$\pm$\footnotesize{0.18}}\\
		\bottomrule
	\end{tabular*}
\end{table}
\begin{table}[t]
	\centering
	\caption{Experimental results compared with other methods on CIFAR-FS and FC100. (The top two performances are shown in bold.)}
	\begin{tabular*}{\hsize}{@{}@{\extracolsep{\fill}}cccccc@{}}
		\toprule
		\label{cifar}
		\multirow{2}{*}{Model} &\multirow{2}{*}{Backbone} 
		&\multicolumn{2}{c}{CIFAR-FS}&\multicolumn{2}{c}{FC100} \\
		\cmidrule{5-6}		\cmidrule{3-4}
		& &5-way 1-shot &5-way 5-shot &5-way 1-shot &5-way 5-shot\\
		\midrule
		Prototypical Net \cite{7}&Conv-64F& 55.50$\pm$\footnotesize{0.70} & 72.00$\pm$\footnotesize{0.60}&35.30$\pm$\footnotesize{0.60}&48.60$\pm$\footnotesize{0.60} \\
		RelationNets \cite{8}&Conv-256F& 55.00$\pm$\footnotesize{1.00} & 69.30$\pm$\footnotesize{0.80}&-&-\\
		R2D2 \cite{23}&Conv-512F& 65.30$\pm$\footnotesize{0.20} & 79.40$\pm$\footnotesize{0.10}&-&-\\
		Prototypical Net \cite{7}&ResNet-12&72.20$\pm$\footnotesize{0.70} & 83.50$\pm$\footnotesize{0.50}& 37.50$\pm$\footnotesize{0.60} & 52.50$\pm$\footnotesize{0.60} \\
		TADAM \cite{24}&ResNet-12& -&-& 40.10$\pm$\footnotesize{0.40}  & 56.10$\pm$\footnotesize{0.40}  \\
		MetaOptNet \cite{29}& ResNet-12&72.60$\pm$\footnotesize{0.70} & 84.30$\pm$\footnotesize{0.50}& 41.10$\pm$\footnotesize{0.60}  & 55.50$\pm$\footnotesize{0.60}  \\
		MABAS \cite{30}& ResNet-12&
		73.51$\pm$\footnotesize{0.92} & 85.49$\pm$\footnotesize{0.68}&
		42.31$\pm$\footnotesize{0.75} & 57.56$\pm$\footnotesize{0.78}  \\
		Fine-tuning \cite{31}&WRN-28-10& \textbf{76.58$\pm$\footnotesize{0.68}} & 85.79$\pm$\footnotesize{0.50}& \textbf{43.16$\pm$\footnotesize{0.59}} & \textbf{57.57$\pm$\footnotesize{0.55}}\\	
		RENet \cite{10} & ResNet12 &74.51$\pm$\footnotesize{0.46}& \textbf{86.60$\pm$\footnotesize{0.32}} &-& - \\													
		\midrule
		SSFormers &ResNet-12&\textbf{74.50$\pm$\footnotesize{0.21}} &\textbf{86.61$\pm$\footnotesize{0.23}}&\textbf{43.72$\pm$\footnotesize{0.21}} & \textbf{58.92$\pm$\footnotesize{0.18}}\\
		\bottomrule
	\end{tabular*}
\end{table}

\section{Experiments}
To evaluate the effectiveness of our method, we conducted extensive experiments on several commonly used benchmarks for few-shot image recognition. In this section, we first present details about datasets and experimental settings in our network design. Then, we compare our method with state-of-the-art methods on various few-shot learning tasks, i.e., standard few-shot learning and semisupervised few-shot learning. Finally, we conduct comprehensive ablation studies to validate each component in our network.

\subsection{Datasets}
We conduct few-shot image recognition problems on four popular benchmarks, i.e., \emph{mini}ImageNet, \emph{tiered}ImageNet, CIFAR-FS and FC100.

\textbf{\emph{mini}ImageNet}~\cite{6} is a subset randomly sampled from ImageNet and is an important benchmark in the few-shot learning community. \emph{mini}ImageNet comprises 60,000 images in 100 categories. We follow the standard partition settings~\cite{7}, where 64/16/20 categories are for training, validation, and evaluation, respectively.

\textbf{\emph{tiered}ImageNet}~\cite{22} is also a subset randomly sampled from ImageNet, which comprises 779,165 images in 608 categories. All 608 categories are grouped into
34 broader categories. Following the same partition settings~\cite{14}, we use 20/6/8 broader categories for training, validation, and evaluation, respectively.

\textbf{CIFAR-FS}~\cite{23} is separated from CIFAR-100, which comprises 60,000 images in 100 categories. CIFAR-FS is divided into 64, 16, and 20 classes for training, validation, and evaluation, respectively.

\textbf{FC100}~\cite{24} is also separated from CIFAR-100, which is more difficult because it is more diverse than CIFAR-FS. FC100 uses a split similar to \emph{tiered}ImageNet, where training, validation, and testing splits contain 60, 20, and 20 classes, respectively.

\subsection{Implementation Details}
\textbf{Backbone networks.} For a fair comparison, following~\cite{25}, we use Conv-64F and ResNet-12 as our model backbone. The Conv-64F~\cite{19,7} contains four repeated blocks. Each block has a convolutional layer with a $3\time 3$ kernel, a batch normalization layer, a ReLU, and a max-pooling with a size of two. We set the number of convolutional channels in each block as 64. In slight contrast to the literature, we add a global max-pooling layer at the end to reduce the dimension of the embedding and obtain patch embeddings. The DropBlock is used in this ResNet architecture to avoid overfitting. In slight contrast to the ResNet-12 in~\cite{29}, we apply a global average pooling after the final layer, which leads to a 640-dimensional patch embedding.

\begin{table}[t]
	\centering
	\caption{The ablation study on our model shows that each part of our model has an important contribution. The experiments are conducted with ResNet12 on \emph{mini}ImageNet. (SSTL: sparse spatial transformer layer, and PMM: patch-matching module.)}
	\begin{tabular*}{\hsize}{@{}@{\extracolsep{\fill}}ccccccc@{}}
			\toprule
			\label{ablation}
			Dense Local Feature&SSTL&SSTL(w/o $m$)&PMM& Cosine Classifier
			&5-way 1-shot & 5-way 5-shot\\
			\midrule	
			&\CheckmarkBold&&\CheckmarkBold&   & 63.15$\pm$\footnotesize{0.20} & 79.15$\pm$\footnotesize{0.25} \\
			\CheckmarkBold&&&\CheckmarkBold&   &\textbf{66.84$\pm$\footnotesize{0.47}}  & 79.72$\pm$\footnotesize{0.50} 	\\
			\CheckmarkBold&\CheckmarkBold&&& \CheckmarkBold  &64.35$\pm$\footnotesize{0.22} & 80.17$\pm$\footnotesize{0.17} \\
			\CheckmarkBold&\CheckmarkBold&\CheckmarkBold&&   	 &66.32$\pm$\footnotesize{0.23} &80.64$\pm$\footnotesize{0.20} \\
			\CheckmarkBold&\CheckmarkBold&&\CheckmarkBold&   	\CheckmarkBold &66.83$\pm$\footnotesize{0.22} &\textbf{83.14$\pm$\footnotesize{0.19}} \\
			\CheckmarkBold&\CheckmarkBold&&\CheckmarkBold&   	 &\textbf{67.25$\pm$\footnotesize{0.24}} &\textbf{82.75$\pm$\footnotesize{0.20}} \\
			\bottomrule
		\end{tabular*}
	\end{table}
	\begin{figure}
		\begin{minipage}{0.5\linewidth}
			\tabcaption{5-way, 1-shot and 5-shot recognition accuracy (\%) with different numbers of image patches on \emph{mini}ImageNet.}
			\begin{tabular}[t]{ccc}
				\toprule
				\label{number}
				Embedding     
				&5-way 1-shot & 5-way 5-shot\\
				\midrule	
				$5\times5$ &65.20$\pm$\footnotesize{0.24} &80.07$\pm$\footnotesize{0.26} \\
				$4\times4$ &66.37$\pm$\footnotesize{0.23} &81.06$\pm$\footnotesize{0.25} \\
				$3\times3$ &65.17$\pm$\footnotesize{0.22} &81.69$\pm$\footnotesize{0.20} \\
				$2\times2$ &65.18$\pm$\footnotesize{0.22} &80.10$\pm$\footnotesize{0.16} \\
				$5\times5+3\times3$ &66.38$\pm$\footnotesize{0.23} &81.82$\pm$\footnotesize{0.24} \\	
				$4\times4+3\times3$ &66.08$\pm$\footnotesize{0.24} &81.50$\pm$\footnotesize{0.26} \\
				$4\times4+2\times2$ &\textbf{67.25$\pm$\footnotesize{0.24}} &\textbf{82.75$\pm$\footnotesize{0.20}} \\	
				$3\times3+2\times2$ &\textbf{67.05$\pm$\footnotesize{0.22}} & \textbf{82.53$\pm$\footnotesize{0.18}} \\
				\bottomrule
			\end{tabular}
			
		\end{minipage}
		\begin{minipage}{0.45\linewidth}
			\centering
			\includegraphics[width=8cm, height=5cm]{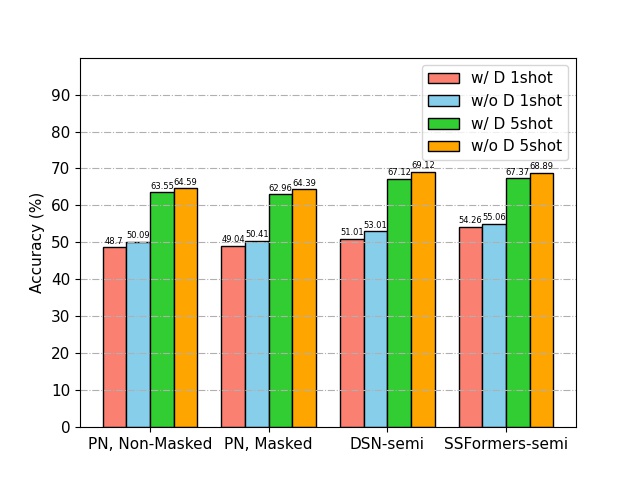} 
			\figcaption{
				5-way semisupervised few-shot learning results on \emph{mini}ImageNet. We show the results with (w/ D) and without (w/o D) \emph{distractors}. Additionally, we compare our methods with PN, Nonmasked~\cite{32}, PN, Masked~\cite{32} and DSN-semi~\cite{25}.}
			\label{fig3}
		\end{minipage}
	\end{figure}
	
	\begin{table}[t]
		\centering
		\caption{5-way, 1-shot and 5-shot recognition accuracy (\%) with different attention methods on \emph{tiered}ImageNet.}
		\begin{tabular*}{\hsize}{@{}@{\extracolsep{\fill}}ccccccc@{}}
			\toprule
			\label{parameter}
			\multirow{2}{*}{Method} &\multirow{2}{*}{Self-att.} 
			&\multirow{2}{*}{Cross-att.}&\multicolumn{2}{c}{\emph{mini}ImageNet} &\multicolumn{2}{c}{\emph{tiered}ImageNet} \\
			&&&1-shot&5-shot&1-shot&5-shot\\
			\midrule	
			Prototypical Net~\cite{7}&&& 62.59$\pm$\footnotesize{0.85} & 78.60$\pm$\footnotesize{0.16}& 68.37$\pm$\footnotesize{0.23} & 83.43$\pm$\footnotesize{0.16}\\
			CAN~\cite{14}&&\CheckmarkBold&  63.85$\pm$\footnotesize{0.48}& 79.44$\pm$\footnotesize{0.34} & 69.89$\pm$\footnotesize{0.51}& 84.23$\pm$\footnotesize{0.37} \\
			FEAT~\cite{19}&&\CheckmarkBold&  66.78$\pm$\footnotesize{0.20}& 82.05$\pm$\footnotesize{0.14}& 70.80$\pm$\footnotesize{0.23}& 84.79$\pm$\footnotesize{0.16}\\
			RENet~\cite{10}&\CheckmarkBold&\CheckmarkBold& \textbf{67.60$\pm$\footnotesize{0.44}} & \textbf{82.58$\pm$\footnotesize{0.30}} & \textbf{71.61$\pm$\footnotesize{0.51}}& \textbf{85.28$\pm$\footnotesize{0.35}}\\	
			SSFormers&&\CheckmarkBold &\textbf{67.25$\pm$\footnotesize{0.24}}  & \textbf{82.75$\pm$\footnotesize{0.20}}  &\textbf{72.52$\pm$\footnotesize{0.25}}  & \textbf{86.61$\pm$\footnotesize{0.18}}\\
			\bottomrule
		\end{tabular*}
	\end{table}
	
	\begin{table}[t]
		\centering
		\caption{Stability evaluation on \emph{mini}ImageNet.}
		\begin{tabular*}{\hsize}{@{}@{\extracolsep{\fill}}ccccc@{}}
			\toprule
			\label{stable}    
			&$+$ GaussianBlur &$+$ PepperNoise &$+$ ColorJitter & $+$ CutMix \\
			\midrule	
			Rethink-D~\cite{40} &82.14$\rightarrow$49.30 &82.14$\rightarrow$63.97 &\textbf{82.14$\rightarrow$81.05}&\textbf{82.14$\rightarrow$71.95}\\ 
			CAN~\cite{14} &79.44$\rightarrow$73.76 &79.44$\rightarrow$65.17 & 79.44$\rightarrow$77.85 &79.44$\rightarrow$70.85\\ 
			CTX~\cite{21} &\textbf{79.54$\rightarrow$75.32}  &\textbf{79.54$\rightarrow$65.85} & 79.54$\rightarrow$78.16 &79.54$\rightarrow$69.35\\ 
			SSFormers &\textbf{82.75$\rightarrow$77.17}&\textbf{82.75$\rightarrow$66.74} &\textbf{82.75$\rightarrow$81.21}&\textbf{82.75$\rightarrow$72.46} \\	 				
			\bottomrule				
			\bottomrule
		\end{tabular*}
	\end{table}
	
	\begin{table*}[h]
		\caption{Comparisons of computation time and FLOPs.}
		\centering
		\begin{tabular*}{\hsize}{@{}@{\extracolsep{\fill}}cccc@{}}
			\toprule
			\label{flops}
			Method  &Time (1-shot)
			&Time (5-shot)
			&FLOPs (G) \\
			\midrule
			DeepEMD~\cite{13}&0.1440&0.0965&-\\
			CAN~\cite{14}&0.0029&0.0035&\textbf{0.0021}\\
			R2D2~\cite{23}&0.0118&0.0044&-\\ 
			RelationNets~\cite{8}&\textbf{0.0008} &\textbf{0.0010}&\textbf{0.0041} \\
			MetaOptNet~\cite{29}&0.3522&1.4018& -\\
			SSFormers&\textbf{0.0022}  &\textbf{0.0027}& \textbf{0.0041} \\
			\bottomrule
			\bottomrule
		\end{tabular*}
	\end{table*}
	
	\begin{figure}[t]
		\centering
		\includegraphics[width=16cm, height=4.4cm]{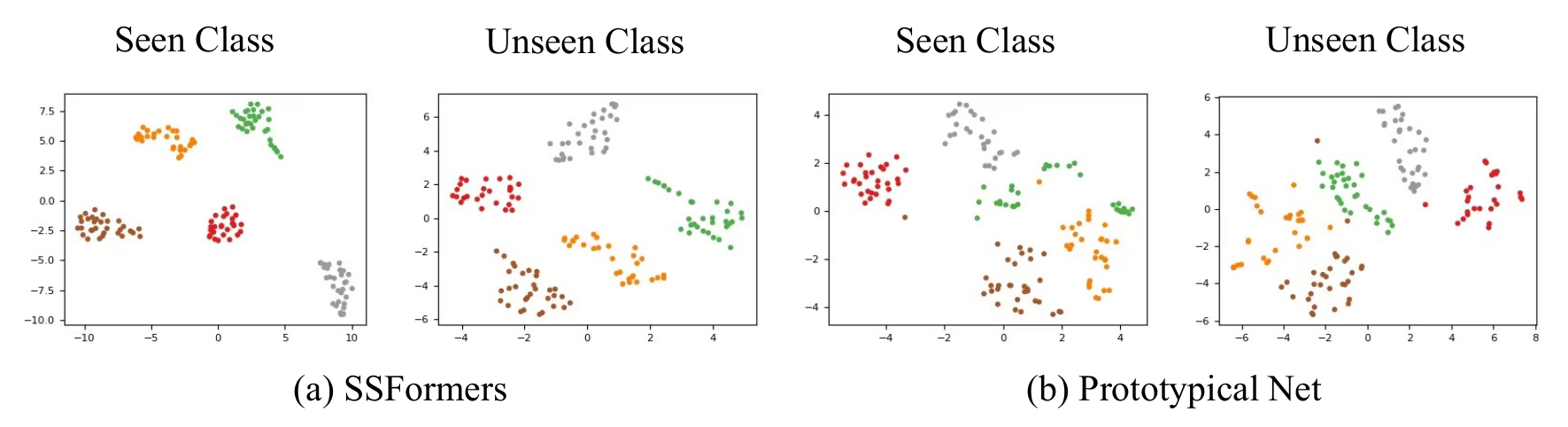} 
		\caption{t-SNE visualization of features for 150 randomly sampled images from five randomly selected classes of the \emph{mini}ImageNet dataset. In our case, the learned embeddings provide better discrimination for seen and unseen classes.}
	\end{figure}
	\begin{figure*}[t]
		\centering
		\includegraphics[width=16cm, height=7.6cm]{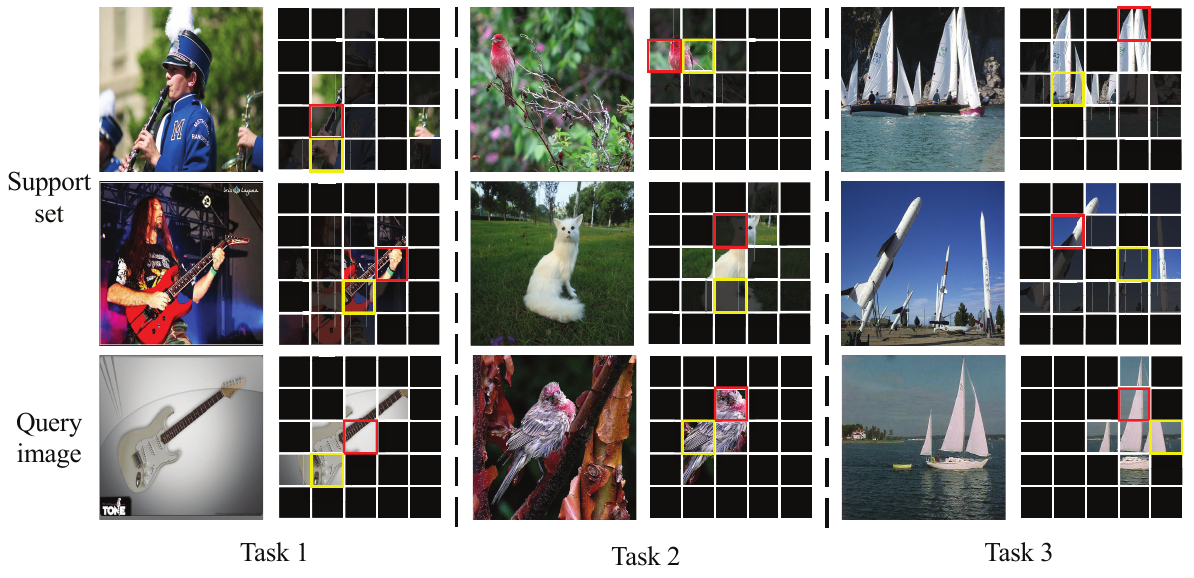} 
		\caption{Visualization of the mask and sparse attention. Given 2-way 1-shot tasks, we plot the masked query and attentioned support images (brighter colors mean higher weight). Within each query image, we choose two image patches (red and yellow boxes) and plot the image patch that best matches each support class. These figures prove that our sparse spatial transformer layer can automatically highlight task-relevant areas.}
	\end{figure*}

\textbf{Patch dividing details.} We select various grids and their combinations (see Table 4). All sampled patches are resized to the input size $84 \times 84$. To generate dense pyramid features, we add a global average pooling layer at the end of the backbone, such that the backbone generates a vector for each input image patch. Moreover, we slightly expand the area of the local patches in the grid twice to merge the context information, which helps generate the local representations. 

\textbf{Training details.} We train Conv-64F from scratch. For ResNet12, the training process can be divided into pre-training and meta-training stages. Following~\cite{13}, we apply a pre-train strategy. The backbone networks are trained on training categories with a softmax layer. In this stage, we apply data argumentation methods to increase the generalization ability of the model, i.e., color jitter, random crop, and random horizontal flip.
The backbone network in our model is initialized with pretrained weights, which are then fine-tuned along with other components in our model. In the meta-training stage, we conduct \emph{N}-way \emph{M}-shot tasks on all benchmarks, i.e., 5-way 1-shot and 5-way 5-shot. Conv-64F is optimized by Adam~\cite{39}, and the initial learning rate is set to 0.1 and decays 0.1 every 10 epochs. Moreover, ResNet-12 is optimized by SGD, and the initial learning rate is set to 5e-4 and decays 0.5 every 10 epochs. We implemented the proposed network using PyTorch~\cite{38}.

\textbf{Evaluation.} During the test stage, we randomly sample 10,000 tasks from the meta-testing set and take the averaged top-1 recognition accuracy as the performance of a method.

\subsection{Standard Few-shot Image Recognition}
To verify the effectiveness of our proposed SSFormers for the few-shot image recognition task, we conduct comprehensive experiments and compare our methods with other state-of-the-art methods. Table 1 and Table 2 list the recognition results of different methods on the \emph{mini}ImageNet, \emph{tiered}ImageNet, CIFAR-FS, and FC100 datasets. The results show that our method achieves the best results in almost all settings. For instance, on the \emph{mini}ImageNet, \emph{tiered}ImageNet, CIFAR-FS, and FC100 datasets, with a ResNet12 backbone, SSFormers have approximately 5.3\%, 3.8\%,3.7\%, and 12.2\% performance improvements, respectively, compared with the vanilla Prototypical Net~\cite{7} under the 5-shot setting, while they have 7.4\%, 6.1\%, 3.2\%, and 16.6\% performance improvements, respectively, under the 1-shot setting.

Note that our model gains 3.3\%/3.2\% and 2.3\%/2.8\% improvements over the most relevant work CTX~\cite{21} under a 1-shot/5-shot setting with ResNet12 on \emph{mini}ImageNet and \emph{tiered}ImageNet, respectively. We obtain this improvement because SSFormers can find task-relevant patches in the current task and perform sparse spatial crossattention algorithms based on dense hierarchical representations.

\subsection{Semi-supervised Few-Shot Learning}
We further verify the effectiveness of our model on more challenging semi-supervised few-shot learning tasks.
Under semisupervised few-shot learning settings, we can select image patches from unlabeled samples that meet the mutual perception function (Eqs. (2)-(4)) and add them to the support set to provide more support features. Particularly, the workflow of SSFormers-semi is as follows. For the support set $n$, we first search for all patches that satisfy the mutual perception function in unlabeled sets and put them into the set $U_n$. Then we use $U_n$ to extend $S_n$:
$S_n = \{S_n^1,...,S_n^{MK}\}\bigcup U_n$. Then, we use the original SSFormers to calculate the similarity.
\label{key}
We use the same experimental setting in~\cite{32}. We use Conv-64F as our backbone and train SSFormers-semi on 300,000 tasks on \emph{mini}ImageNet with 40\% labeled data. The results are shown in Figure 3, where SSFormers-semi shows competitive results with classical baseline methods.

\subsection{Ablation Study}
\textbf{Analysis of our method.}
Our model comprises different components: a dense local feature extractor, sparse spatial transformer layer (SSTL), and PMM. As shown in Table 3, we verify the indispensability of each component on the \emph{mini}ImageNet dataset. Note that we calculate the cosine similarity spatially when replacing the PMM. The results show that every component in SSFormers has an important contribution. For example, without our SSTL, the performance of the model drops by 3.73\% on 5-shot tasks. Additionally, if we do not use sparse mask $m$, our model is equivalent to using transformers for crossattention between two images. As shown in Table 3, performance is reduced by 1.38\%/2.55\% for 1-shot/5-shot tasks, respectively. Moreover, if we introduce an additional cosine classifier for SSFormers and dynamically adjust the weights of the PMM and cosine classifier through a learnable weight parameter, the performance of the model is slightly reduced for 1-shot tasks and slightly improved for 5-shot tasks. These results are obtained because the cosine distance for global-level features is not very effective when the samples are scarce, as verified by the results in~\cite{7,44}.

\textbf{Influence of the number of patches.}
While dividing input images into patches, we must define the grid for patches. We select various grids and combinations to conduct analysis experiments on \emph{mini}ImageNet. As shown in Table 4, an approach is to use a combination of grids of different sizes. A possible explanation is that the size of the main object differs between images and using a single size may lose context information and make high-level semantic representations difficult to generate. 

\textbf{Comparison with other attention methods.}
As shown in Table 5, our model achieves competitive results with other state-of-the-art attention-based methods. Particularly, CANs propose a crossattention algorithm to highlight the common objects in an image pair. FEAT uses a transformer to model support image relationships and generate task-specific support features. FEAT neglects query image information, which could result in information loss. RENet uses 4-dimensional convolution for self-attention and crossattention between two images. Our SSFormers consider not only the relationship between two images but also the relationship between the query image and all support images in the current task. SSFormers select task-relevant patches in the query image to align the support set with the query image using a sparse spatial crossattention algorithm.

\textbf{Analyze stability.}
A good model should have good robustness and adaptability to various environments. For this reason, we tested the stability of our model under three attacks on \emph{mini}ImageNet. As shown in Table 6, the performance of our SSFormers is relatively stable under various attacks, i.e., GaussianBlur ($\delta\in\left[0.1, 2\right]$), PepperNoise ($r=0.01$), ColorJitter ($B=0.8$), and CutMix~\cite{45}. For example, when this model was attacked by GaussianBlur, the performance decreased by 40.0\% for Rethink-D~\cite{40} but only by 6.7\% for SSFormers.

\textbf{Time Complexity.} We compare our SSFormers with other state-of-the-art methods regarding time complexity. The computation time is averaged by $10^4$ forward inferences, and each task only contains one query image. SSFormers are observed to have comparably little computation time compared to a CAN~\cite{14}. However, a CAN~\cite{14} has lower FLOPs because the partitioning of patches and feature extracting brings additional computation.

\textbf{Qualitative visualizations.} 
We do a t-SNE visualization of the output embeddings from SSFormers from the seen and unseen images of \emph{mini}ImageNet to demonstrate the effectiveness of our method (see Figure 4). Our method maintains good class discrimination compared to Prototypical Net, even for unseen test classes. Moreover, the features generated by our method are more discriminative, and the boundaries between categories are more prominent than that in Prototypical Net. We provide visualization cases in Figure 5 to further qualitatively evaluate the proposed SSFormers. For each query image in the task, we plot the result of its mask, and it is seen that through the mutual perception function (Eqs. (2)-(4)), we can obtain task-relevant query image patches. For support sets, we plot the prototype generated after SSTL. Our model is shown to highlight task-relevant image patches and suppress task-irrelevant features.
 
\section{Conclusion}
In this article, we argue that global features and deep descriptors are ineffective for few-shot learning because global features lose local information, and deep descriptors lose the contextual information of images.
Moreover, a common embedding space fails to generate discriminative visual representations for a target task. On the basis of this fact, we propose novel sparse spatial transformers (SSFormers) to help meta-classifiers learn more discriminative features. In SSFormers, we use patch-level features for representation and propose a novel SSTL to customize task-specific prototypes through a transformer-based architecture. We propose a nonparametric PMM to obtain patch-level similarity for final recognition. Experimental results demonstrate that SSFormers can achieve competitive results with other state-of-the-art few-shot learning methods. In the future, we will explore designing a pure vision transformer for few-shot learning, which can substantially reduce the time cost of the patch dividing step and fully exploit the powerful learning representation of the vision transformer.

\Acknowledgements{This work was partially supported by the National Natural Science Foundation of China (Nos. 62176116, 62073160, 62276136), and the Natural Science Foundation of the Jiangsu Higher Education Institutions of China, No. 20KJA520006.}

\end{document}